\documentclass[a4paper,fleqn]{cas-dc}

\usepackage[authoryear]{natbib}

\usepackage{amssymb}
\usepackage{amsmath}
\usepackage{tabularx}
\usepackage{adjustbox}
\usepackage{graphicx}   
\usepackage{multirow}   
\usepackage{array}
\usepackage{makecell}
\usepackage{booktabs}

\newcommand{\ourmethod}{ForestQuery}

\begin{document}
\let\WriteBookmarks\relax
\def\floatpagepagefraction{1}
\def\textpagefraction{.001}
\shorttitle{ForestQuery}
\shortauthors{Zhan et~al.}

\title [mode = title]{ForestQuery: Boundary-Aware and Spatially Anchored Query Learning for Unified Forest Point Cloud Segmentation}                      

\author[a,b]{Zhihao Zhan}[orcid=0009-0003-9065-5261]

\author[b]{Le Tao}[orcid=0009-0004-7566-6584]

\author[c]{Yifei Tian}[orcid=0000-0001-8884-2170]

\author[d]{Xin Liu}[orcid=0000-0002-7658-284X]

\author[a]{Jie Yuan}[orcid=0000-0003-0597-7657]
\cormark[1]
\cortext[cor]{Jie Yuan (yuanjie@nju.edu.cn) is the corresponding author.}

\affiliation[a]{
            organization={School of Electronic Science and Engineering, Nanjing University},
            city={Nanjing},
            postcode={210023}, 
            country={China}}
\affiliation[b]{
            organization={TopXGun Robotics},
            city={Nanjing},
            postcode={211100}, 
            country={China}}
\affiliation[c]{
            organization={School of Computer Science, Nanjing University of Posts and Telecommunications},
            city={Nanjing},
            postcode={210023}, 
            country={China}}
\affiliation[d]{
            organization={Department of Computer Science and Technology, Tsinghua University},
            city={Beijing},
            postcode={100084}, 
            country={China}}

\begin{abstract}
Forest point cloud segmentation is fundamental for fine-grained 3D forest scene understanding, yet remains challenging due to irregular tree structures, severe occlusions, density variations, and ambiguous instance boundaries.
Recent query-based forest segmentation methods have shown promise for unified semantic and instance prediction, but they still insufficiently exploit forest-specific spatial structure and account for boundary uncertainty.
In this paper, we propose \ourmethod, a boundary-aware and spatially anchored query learning framework for unified forest point cloud segmentation.
\ourmethod~enhances instance and semantic query learning through two complementary designs.
Specifically, boundary uncertainty is explicitly modeled to guide reliable instance query construction and modulate query optimization through adaptive loss reweighting.
Meanwhile, spatially anchored semantic query enhancement (SA-SQE) introduces learnable 3D anchors encoding forest vertical stratification priors to enrich semantic queries with explicit spatial references.
We evaluate \ourmethod~on multiple public forest point cloud benchmarks and a self-collected annotated real-world dataset.
Extensive experiments demonstrate consistent improvements in both individual-tree segmentation and semantic segmentation across diverse forest scenes.
Code and data are publicly available at \url{https://zhan994.github.io/ForestQuery}.
\end{abstract}







\maketitle

\section{Introduction}
\label{sec:intro}
Understanding complex 3D natural scenes is a fundamental problem in computer vision.
Recent advances in 3D sensing and scene reconstruction have enabled large-scale outdoor environments to be represented as point clouds with rich geometric and structural information~\citep{liu2021individual,wilkes2023tls2trees,zhan2025agrilira4d,zhan2026terrain}.
Forest point clouds are particularly challenging due to irregular spatial layouts, heterogeneous tree structures, severe occlusions, and ambiguous object boundaries.
This work focuses on unified forest point cloud segmentation, which jointly performs point-wise semantic labeling of forest components and instance-level separation of individual trees.
Such fine-grained forest understanding provides essential support for forest inventory, ecological monitoring, and sustainable forest management~\citep{waser2021mapping,xiang2024automated,xiang2025forestformer3d,tan2026tlnet}.

General point cloud segmentation has been studied with a variety of representations, including point-based, voxel-based, sparse-convolutional, and Transformer-based architectures~\citep{qi2017pointnet,qi2017pointnetplusplus,tchapmi2017segcloud,choy20194d,zhao2021point}.
For instance segmentation, existing approaches can be broadly categorized into top-down, bottom-up, and query-based paradigms.
Top-down methods first generate object proposals or bounding boxes and subsequently predict instance masks within individual proposals~\citep{yi2019gspn,hou20193d}, whereas bottom-up methods typically predict point-wise embeddings, offsets, or affinities and assemble points into instances through clustering or hierarchical grouping~\citep{jiang2020pointgroup,chen2021hierarchical,vu2022softgroup}.
These approaches often rely on accurate proposal generation or carefully designed grouping and post-processing strategies.
More recently, query-based Transformer methods directly decode segmentation masks through interactions between learnable queries and scene features~\citep{schult2023mask3d,kolodiazhnyi2024oneformer3d}.
This formulation reduces reliance on explicit proposal generation and hand-crafted grouping, while naturally supporting unified semantic and instance segmentation.
The query-based paradigm has also been extended to forest point clouds for joint semantic labeling and individual-tree segmentation~\citep{xiang2025forestformer3d}.

\begin{figure*}[htbp]
    \centering
    \includegraphics[width=0.95\linewidth]{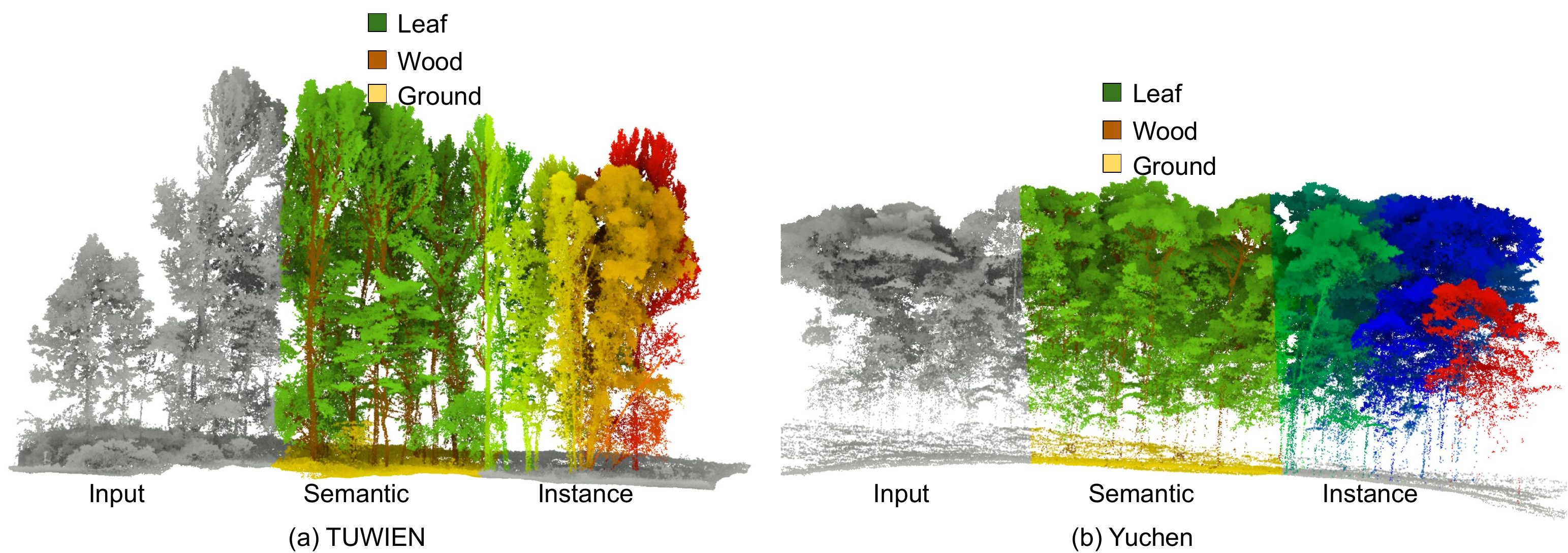}
    \caption{\textbf{Representative examples of unified forest point cloud segmentation.}
    Examples from the TUWIEN and Yuchen scenes of FOR-instanceV2 illustrate the complex spatial organization of forest point clouds, where heterogeneous vegetation structures and overlapping tree crowns pose challenges to semantic understanding and individual-tree separation.}
    \label{fig:challenge}
\end{figure*}

As illustrated by the representative scenes in Fig.~\ref{fig:challenge}, unified forest point cloud segmentation must cope with complex spatial organization, heterogeneous vegetation structures, and closely interacting tree crowns.
Despite recent progress, existing query-based formulations remain insufficiently tailored to these forest-specific characteristics.
For instance segmentation, severe crown overlap and intertwined branches create ambiguous transition regions between neighboring trees, where points may contain mixed instance information.
Queries initialized from such regions can therefore yield unreliable instance representations, while uniform supervision may further amplify the influence of boundary uncertainty during optimization.
For semantic segmentation, forest scenes exhibit pronounced vertical stratification, with different semantic components distributed across distinct yet partially overlapping spatial layers.
Existing semantic queries, however, are typically initialized without explicit spatial references and must recover these structural regularities solely through interactions with scene features.
Such insufficient spatial anchoring makes semantic discrimination particularly difficult for components with similar local geometry.
These observations motivate forest-specific query learning that explicitly accounts for boundary uncertainty while incorporating spatial structure into semantic representations.

To address these challenges, we propose \ourmethod, a boundary-aware and spatially anchored query learning framework for unified forest point cloud segmentation.
For instance segmentation, boundary uncertainty guides reliable query construction under ambiguous inter-tree boundaries.
In parallel, spatially anchored semantic query enhancement (SA-SQE) incorporates learnable 3D anchors encoding forest vertical stratification priors into semantic queries.
The resulting queries are jointly decoded with shared scene features, with boundary uncertainty further used to adaptively reweight instance-query supervision.
Extensive experiments on multiple public benchmarks and a self-collected annotated real-world dataset demonstrate the effectiveness and cross-scene generalization of \ourmethod.

The main contributions are summarized as follows:
\begin{itemize}
    \item We propose \ourmethod, a unified query-based framework for joint individual-tree instance segmentation and point-wise semantic labeling in complex forest scenes.

    \item We develop a boundary-aware mechanism that explicitly models boundary uncertainty and exploits it in both instance query construction and optimization, enabling more reliable separation of densely overlapping trees.
    
    \item We introduce SA-SQE, which incorporates learnable 3D anchors encoding forest vertical stratification priors, enriching semantic queries with explicit spatial references.

    \item We evaluate \ourmethod~on multiple public benchmarks and a self-collected annotated real-world dataset, covering diverse forest scenes and acquisition conditions.
\end{itemize}

The remainder of this paper is organized as follows.
Section~\ref{sec:related} reviews related work on general point cloud segmentation and forest scene segmentation.
Section~\ref{sec:method} presents \ourmethod, covering boundary-aware instance query construction, spatially anchored semantic query enhancement, and unified query decoding and optimization.
Section~\ref{sec:exp} reports the experimental setup, comparative evaluations, ablation studies, and generalization tests.
Finally, Section~\ref{sec:conclusion} concludes the paper and outlines future work..

\section{Related Work}
\label{sec:related}

\subsection{Point Cloud Segmentation}
\label{sec:related_segment}

Point cloud segmentation is a fundamental task in 3D scene understanding.
Early deep networks, represented by PointNet~\citep{qi2017pointnet} and PointNet++~\citep{qi2017pointnetplusplus}, established a direct point-based paradigm by learning representations from unordered point sets.
Building on this foundation, subsequent methods strengthened local geometric modeling through convolutional operators defined over points or voxelized space.
KPConv~\citep{thomas2019kpconv} performs kernel point convolution over local neighborhoods, whereas SEGCloud~\citep{tchapmi2017segcloud} adopts a volumetric representation and projects predictions back to raw points.
To improve scalability in large-scale scenes, sparse-convolutional architectures exploit the sparsity of 3D data for efficient feature extraction~\citep{choy20194d}, while RandLA-Net~\citep{hu2020randla} combines random sampling with local feature aggregation for efficient point-wise prediction.
More recently, Transformer-based models further extend point cloud representation learning by capturing long-range contextual dependencies.
Point Transformer~\citep{zhao2021point} introduces self-attention tailored to irregular point sets, Stratified Transformer~\citep{lai2022stratified} enlarges the effective receptive field through stratified key sampling, and Point Transformer V3 (PTv3)~\citep{wu2024ptv3} further improves efficiency and scalability through serialized point-cloud processing.
Collectively, these advances have substantially strengthened point-wise semantic understanding, while fine-grained scene interpretation further requires distinguishing individual object instances.

Instance segmentation extends point-wise recognition toward object-level scene decomposition.
Existing methods can be broadly categorized into top-down, bottom-up, and query-based paradigms.
Top-down approaches first localize candidate objects and subsequently predict instance masks within individual proposals.
Representative methods include GSPN~\citep{yi2019gspn}, which generates object proposals directly from point clouds, and 3D-SIS~\citep{hou20193d}, which jointly performs 3D object localization and instance mask prediction.
Bottom-up methods instead construct object instances from point-wise predictions.
SGPN~\citep{wang2018sgpn} learns pairwise similarities for instance grouping, while PointGroup~\citep{jiang2020pointgroup} and HAIS~\citep{chen2021hierarchical} organize points into instances through offset-based grouping and hierarchical aggregation, respectively.
SoftGroup~\citep{vu2022softgroup} further adopts soft semantic grouping followed by proposal refinement to improve instance formation.
Although effective, top-down and bottom-up methods often depend on proposal generation, clustering, or carefully designed grouping procedures.
More recently, query-based methods directly represent object instances with learnable queries and decode their masks through interactions with scene features.
Mask3D~\citep{schult2023mask3d} employs Transformer decoders to iteratively associate instance queries with multi-scale point features, while QueryFormer~\citep{lu2023query} further investigates query initialization and refinement for robust instance prediction.
Building on this paradigm, OneFormer3D~\citep{kolodiazhnyi2024oneformer3d} unifies semantic, instance, and panoptic segmentation within a shared query-based framework.
These advances establish a natural foundation for extending unified query-based segmentation to complex forest scenes.

\subsection{3D Forest Scene Segmentation}
\label{sec:related_forest}

Forest point cloud segmentation has been extensively studied for recovering tree- and component-level structures from 3D observations.
Early approaches mainly relied on geometric priors and hand-crafted procedures, including canopy height models, region growing, clustering, and rule-based tree detection~\citep{naveed2019individual,liu2021individual,wilkes2023tls2trees}.
Although effective in relatively regular forest stands, such methods often degrade under severe occlusion, heterogeneous point densities, understory vegetation, and overlapping crowns.
Learning-based approaches alleviate these limitations by learning forest representations directly from point clouds.
ForAINet~\citep{xiang2024automated} jointly predicts semantic labels and individual-tree instances for automated forest inventory, while SegmentAnyTree~\citep{wielgosz2024segmentanytree} adopts a PointGroup-inspired architecture to improve transferability across sensing platforms and point densities.
TreeLearn~\citep{henrich2024treelearn} further develops a data-driven pipeline for individual-tree segmentation from ground-based forest point clouds.
Despite their effectiveness, these methods largely retain grouping- or clustering-based instance formation, making accurate tree separation difficult in scenes with complex canopy interactions and large structural variations.

Building on these learning-based formulations, recent studies have increasingly incorporated semantic, structural, and interactive cues into individual-tree segmentation.
3DPS-Net~\citep{xiu2025individual} introduces prompt-driven mask prediction for interactive and automatic tree extraction.
\cite{ma2025individual} incorporate semantic priors and contrastive learning into instance clustering to improve tree separation under canopy overlap.
SATree~\citep{du2026satree} jointly models semantic categories, stem locations, and point-wise offsets to facilitate structure-aware instance grouping, whereas ForestSeg3D~\citep{zhang2026forestseg3d} combines hierarchical semantic supervision with cross-task distillation to improve tree delineation in crowded forest scenes.
These studies demonstrate the benefit of introducing semantic and structural cues into instance reasoning.
Nevertheless, their primary objective remains individual tree extraction, and instance formation still commonly relies on explicit grouping, clustering, or proposal refinement.

The demand for joint scene-component understanding and individual-tree delineation has further motivated unified forest segmentation frameworks.
ForestFormer3D~\citep{xiang2025forestformer3d} introduces an end-to-end query-based formulation that jointly predicts point-wise semantic labels and individual-tree instances, incorporating instance-aware query selection and one-to-many association for forest-specific decoding.
SegmentAnyTreeV2~\citep{wielgosz2026segmentanytreev2} combines a PTv3~\citep{wu2024ptv3} backbone with cross-attention decoding and instance-aware query initialization to improve semantic and instance prediction across diverse forest conditions.
ForestMamba~\citep{nguyen2026forestmamba} incorporates forest-specific geometric cues at both the representation and query levels, using vertical-priority voxel serialization for Mamba encoding and canopy-height-guided initialization for tree queries.
Beyond query-based formulations, CTS~\citep{xie2026cts} adopts a panoptic formulation that jointly models semantic tree components and instance identities through semantic grouping and feature clustering.
Collectively, these studies highlight the potential of unified semantic-instance reasoning for fine-grained forest scene understanding.
However, existing query-based formulations provide limited mechanisms for explicitly modeling boundary uncertainty during instance query construction and optimization, as well as for incorporating forest-specific spatial structure into semantic query representations.
Our work addresses these complementary limitations through a boundary-aware mechanism for instance queries and spatially anchored semantic query enhancement.

\section{Methodology}
\label{sec:method}

\begin{figure*}[htbp]
    \center
    \includegraphics[width=\linewidth]{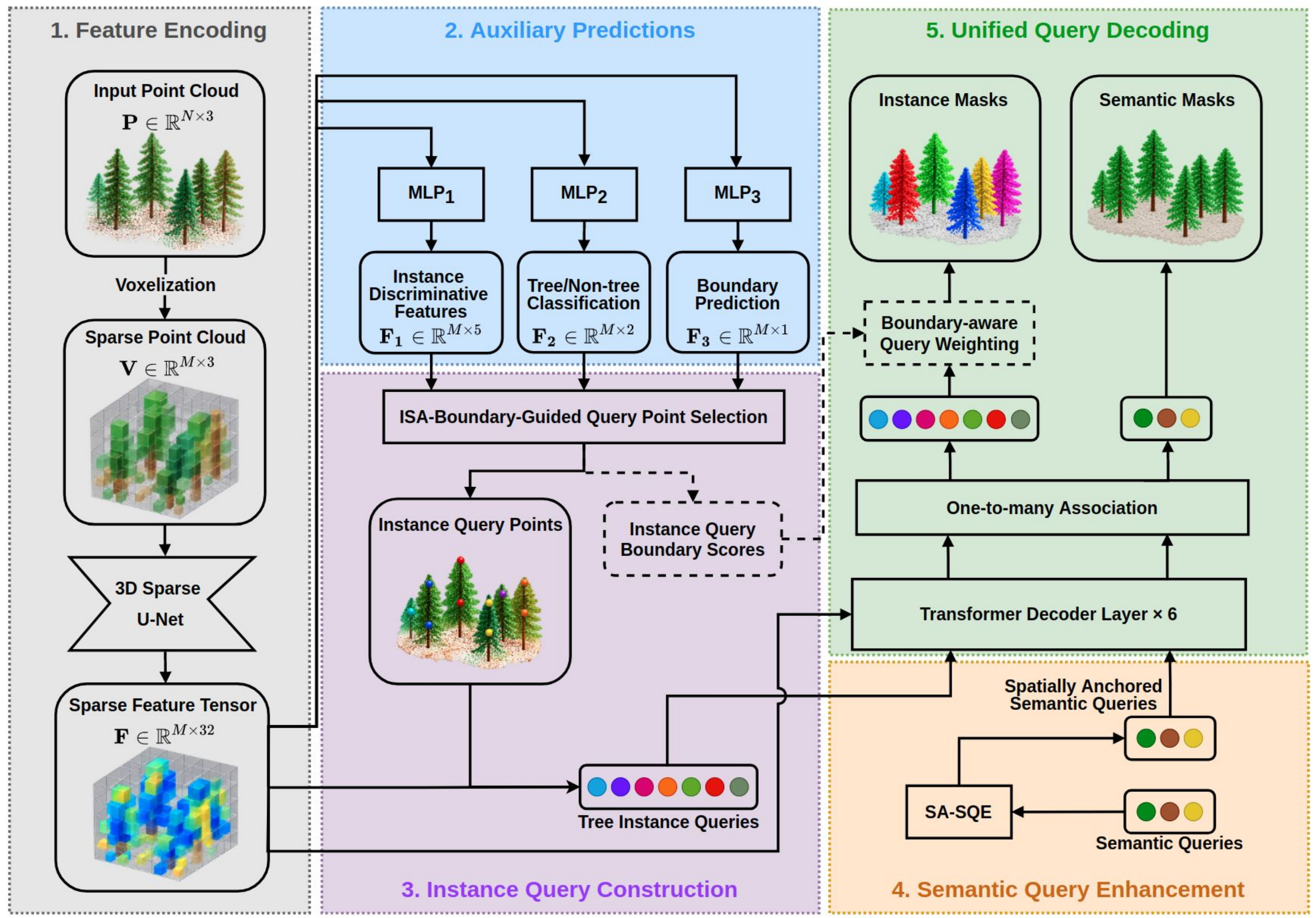}
    \caption{\textbf{Overview of the proposed \ourmethod~framework.}  
    A shared sparse 3D backbone supports boundary-aware instance query construction and joint query decoding.
    Boundary cues guide instance query construction and optimization, while SA-SQE injects learnable 3D anchors encoding forest vertical stratification priors into semantic queries.
    The resulting queries are jointly decoded with shared scene features to predict semantic and individual-tree instance masks.
    }
    \label{fig:framework}
\end{figure*} 

\subsection{Problem Formulation and Overview}
\label{sec:method_overview}

Given a forest point cloud $\mathbf{P}\in\mathbb{R}^{N\times3}$ containing $N$ points, where $\mathbf{p}_i=[x_i,y_i,z_i]^{\top}\in\mathbb{R}^{3}$ denotes the $i$-th point, we formulate unified forest point cloud segmentation as the joint prediction of semantic labels and individual-tree instance identities.
For each point $\mathbf{p}_i$, the model predicts a semantic label $y_i^{\mathrm{sem}}\in\{1,\ldots,C\}$ and, for tree points, an instance label $y_i^{\mathrm{ins}}\in\{1,\ldots,K\}$, where $C$ is the number of semantic categories, and $K$ denotes the number of tree instances.
Non-tree points are excluded from tree-instance assignment.

As illustrated in Fig.~\ref{fig:framework}, \ourmethod~adopts a query-based formulation built upon shared sparse 3D features.
The input point cloud is first voxelized and encoded by a sparse 3D backbone to obtain a feature tensor $\mathbf{F}\in\mathbb{R}^{M\times d}$, where $M$ denotes the number of active voxels and $d$ is the feature dimension.
Instance and semantic queries are then constructed through two complementary pathways.
For instance queries, instance-discriminative features, coarse semantic cues, and boundary cues derived from $\mathbf{F}$ are integrated through the proposed instance-semantic-aware (ISA)-boundary-guided query point selection strategy.
In parallel, SA-SQE enhances a set of learnable semantic queries with learnable 3D anchors initialized according to forest vertical stratification, thereby introducing explicit spatial priors.

Let $\mathbf{Q}^{\mathrm{ins}}$ and $\widetilde{\mathbf{Q}}^{\mathrm{sem}}$ denote the resulting instance and spatially anchored semantic queries, respectively.
They are jointly processed by a Transformer decoder, with the shared sparse feature tensor $\mathbf{F}$ serving as scene memory:
\begin{equation}
    \left(
    \mathbf{Q}^{\mathrm{ins}}_{L},
    \mathbf{Q}^{\mathrm{sem}}_{L}
    \right)
    =
    \operatorname{Dec}^{(L)}
    \left(
    \mathbf{Q}^{\mathrm{ins}},
    \widetilde{\mathbf{Q}}^{\mathrm{sem}},
    \mathbf{F}
    \right),
\end{equation}
where $\operatorname{Dec}^{(L)}$ denotes an $L$-layer query decoder.
The decoded queries are subsequently mapped to individual-tree instance masks and semantic masks.
During optimization, the predicted boundary uncertainty further modulates instance-query supervision through adaptive loss reweighting.
The following sections detail instance query construction, semantic query enhancement, and unified query decoding and optimization.

\subsection{Sparse Feature Encoding and Auxiliary Predictions}
\label{sec:method_feature_query}

Given the input point cloud $\mathbf{P}$, we first voxelize the irregular points into a sparse voxel representation as $\mathbf{V} = \mathcal{V}(\mathbf{P}) \in\mathbb{R}^{M\times3}$, where $\mathcal{V}(\cdot)$ denotes the voxelization operation and $M$ is the number of active voxels, typically much smaller than $N$.
A 3D sparse U-Net~\citep{choy20194d} then encodes $\mathbf{V}$ to obtain the shared feature representation as $ \mathbf{F} = \mathcal{E}(\mathbf{V}) \in\mathbb{R}^{M\times d}$, where $\mathcal{E}(\cdot)$ denotes the sparse feature encoder and $d=32$ is the feature dimension.
The resulting feature tensor $\mathbf{F}$ serves as the shared scene representation for auxiliary prediction, instance query construction, and subsequent query decoding.

To provide complementary cues for query point selection, three auxiliary prediction heads operate on the shared voxel features.
The instance-discriminative and coarse semantic heads are defined as
\begin{equation}
    \mathbf{F}_{1}=\phi_{1}(\mathbf{F})\in\mathbb{R}^{M\times5},
    \quad
    \mathbf{F}_{2}=\phi_{2}(\mathbf{F})\in\mathbb{R}^{M\times2},
\end{equation}
where $\mathbf{F}_{1}$ denotes the instance-discriminative features and $\mathbf{F}_{2}$ denotes the coarse tree/non-tree log-probability scores.
A dedicated boundary head predicts boundary logits from the shared features:
\begin{equation}
    \mathbf{F}_{3}
    =
    \phi_{3}(\mathbf{F})
    \in\mathbb{R}^{M\times1}.
\end{equation}
These complementary outputs are subsequently exploited by the ISA-boundary-guided query point selection strategy.

\subsection{Boundary-Aware Instance Query Construction}
\label{sec:method_boundary_query}

\subsubsection{Boundary Target Construction and Supervision}

\begin{figure}
    \center
    \includegraphics[width=0.95\linewidth]{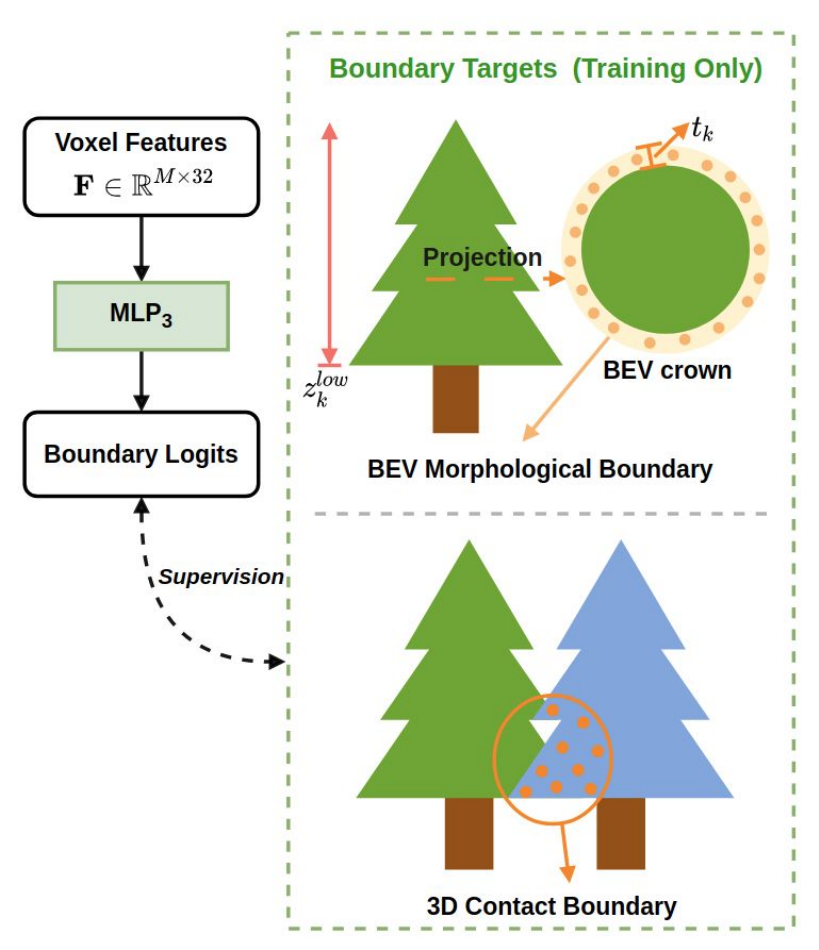}
    \caption{\textbf{Instance boundary target construction and prediction.}
    Boundary targets are constructed from two complementary cues: adaptive BEV morphological boundaries characterize projected crown contours, while 3D contact boundaries identify locally adjacent regions belonging to different tree instances.
    A lightweight boundary head predicts boundary logits from the shared voxel features under the constructed supervision.
    }
    \label{fig:boundary_branch}
\end{figure} 

As illustrated in Fig.~\ref{fig:boundary_branch}, we construct boundary targets from two cues.
The bird's-eye-view (BEV) morphological boundary captures the projected crown boundary of each tree, whereas the 3D contact boundary identifies local regions where neighboring tree instances are spatially adjacent.
Let $\mathcal{S}=\{v_i\}_{i=1}^{M}$ denote the set of active voxels in the voxelized representation $\mathbf{V}$, where each voxel $v_i$ is associated with its center coordinate $\mathbf{c}_i=[x_i,y_i,z_i]^{\top}$.
Let $\mathcal{S}_k\subseteq\mathcal{S}$ denote the voxels belonging to the $k$-th tree instance.

For each instance, we first retain its upper-canopy voxels and project them onto the BEV plane:
\begin{equation}
    \mathcal{S}_k^{\mathrm{bev}}
    =
    \left\{
    (x_i,y_i)
    \;\middle|\;
    v_i\in\mathcal{S}_k,\;
    z_i>z_k^{\mathrm{low}}
    \right\}.
\end{equation}

The lower canopy threshold is defined as
\begin{equation}
    z_k^{\mathrm{low}}
    =
    z_k^{\min}
    +
    \eta
    \left(
    z_k^{\max}-z_k^{\min}
    \right),
\end{equation}
where $z_k^{\min}$ and $z_k^{\max}$ denote the minimum and maximum heights of the $k$-th tree instance, respectively, and $\eta$ controls the relative height used to separate the upper-canopy region.

To adapt the boundary width to crown size, we estimate the projected crown area, equivalent radius, and erosion thickness as
\begin{equation}
    A_k=
    \left|
    \mathcal{S}_k^{\mathrm{bev}}
    \right|s_v^2,
    \quad
    r_k=\sqrt{\frac{A_k}{\pi}},
    \quad
    t_k=
    \operatorname{round}
    \left(
    \frac{\alpha r_k}{s_v}
    \right),
\end{equation}
where $s_v$ denotes the BEV grid resolution and $\alpha$ controls the relative boundary width.
Applying morphological erosion yields $\mathcal{S}_k^{\mathrm{bev,ero}}$, and the BEV morphological boundary is defined as
\begin{equation}
    \mathcal{R}_k^{\mathrm{bev}}
    =
    \mathcal{S}_k^{\mathrm{bev}}
    \setminus
    \mathcal{S}_k^{\mathrm{bev,ero}}.
\end{equation}
The resulting boundary is mapped back to the voxel domain:
\begin{equation}
    \mathcal{B}_k^{\mathrm{bev}}
    =
    \left\{
    v_i\in\mathcal{S}_k
    \;\middle|\;
    (x_i,y_i)\in\mathcal{R}_k^{\mathrm{bev}}
    \right\}.
\end{equation}

The BEV construction captures crown morphology but does not explicitly model local interactions between adjacent trees.
We therefore additionally construct a 3D contact boundary.
Let
\begin{equation}
    \mathcal{S}_{\mathrm{tree}}
    =
    \left\{
    v_i\in\mathcal{S}
    \;\middle|\;
    y_i^{\mathrm{bi}}=1
    \right\},
\end{equation}
where $y_i^{\mathrm{bi}}\in\{0,1\}$ denotes the ground-truth tree/non-tree label of voxel $v_i$.
For a tree voxel $v_i$, its local 3D neighborhood is defined as
\begin{equation}
    \mathcal{N}(v_i)
    =
    \left\{
    v_j\in\mathcal{S}
    \;\middle|\;
    \|\mathbf{c}_j-\mathbf{c}_i\|_2
    \le r_{\mathrm{c}}
    \right\},
\end{equation}
where $r_{\mathrm{c}}$ denotes the neighborhood radius.
A tree voxel is regarded as a contact-boundary voxel if at least one neighboring tree voxel belongs to a different instance:
\begin{equation}
\begin{aligned}
    \mathcal{B}^{\mathrm{contact}}
    =
    \Big\{
    v\in\mathcal{S}_{\mathrm{tree}}
    \;\Big|\;
    &\exists\,v'\in
    \mathcal{N}(v)\cap\mathcal{S}_{\mathrm{tree}},\\
    &y^{\mathrm{ins}}(v)
    \neq
    y^{\mathrm{ins}}(v')
    \Big\}.
\end{aligned}
\end{equation}

The complete boundary set is obtained by combining both cues:
\begin{equation}
    \mathcal{B}
    =
    \left(
    \bigcup_{k=1}^{K}
    \mathcal{B}_k^{\mathrm{bev}}
    \right)
    \cup
    \mathcal{B}^{\mathrm{contact}},
\end{equation}
and the corresponding binary boundary target is set to $b_i^{\mathrm{gt}}=1$ for $v_i\in\mathcal{B}$ and $b_i^{\mathrm{gt}}=0$ otherwise.

To obtain explicit boundary cues, a lightweight boundary head is applied to the shared voxel features.
For voxel $v_i$ with feature $\mathbf{f}_i\in\mathbb{R}^{d}$, the corresponding boundary logit is $\ell_i^{\mathrm{bd}} = \phi_{3}\left( \mathbf{f}_i \right)$, where $\phi_{3}(\cdot)$ denotes the boundary prediction MLP.
The corresponding boundary probability is $b_i = \sigma \left( \ell_i^{\mathrm{bd}}\right)$, where $\sigma(\cdot)$ denotes the sigmoid function.
Boundary supervision is applied only to tree locations, while positive boundary samples are assigned a larger weight to alleviate class imbalance.
The boundary loss is defined as
\begin{equation}
\begin{aligned}
    \mathcal{L}_{\mathrm{bd}}
    =
    -\frac{1}{M}
    \sum_{i=1}^{M}
    y_i^{\mathrm{bi}}
    \Big[
        &w_{\mathrm{pos}} b_i^{\mathrm{gt}}\log b_i \\
        &+
        (1-b_i^{\mathrm{gt}})\log(1-b_i)
    \Big],
\end{aligned}
\end{equation}
where $w_{\mathrm{pos}}>1$ denotes the positive-class weight.
In implementation, the loss is computed directly from the boundary logits using the numerically stable binary-cross-entropy-with-logits formulation.
The resulting boundary probabilities provide explicit boundary cues for the subsequent ISA-boundary-guided query point selection.

\begin{figure*}[htbp]
    \center
    \includegraphics[width=0.95\linewidth]{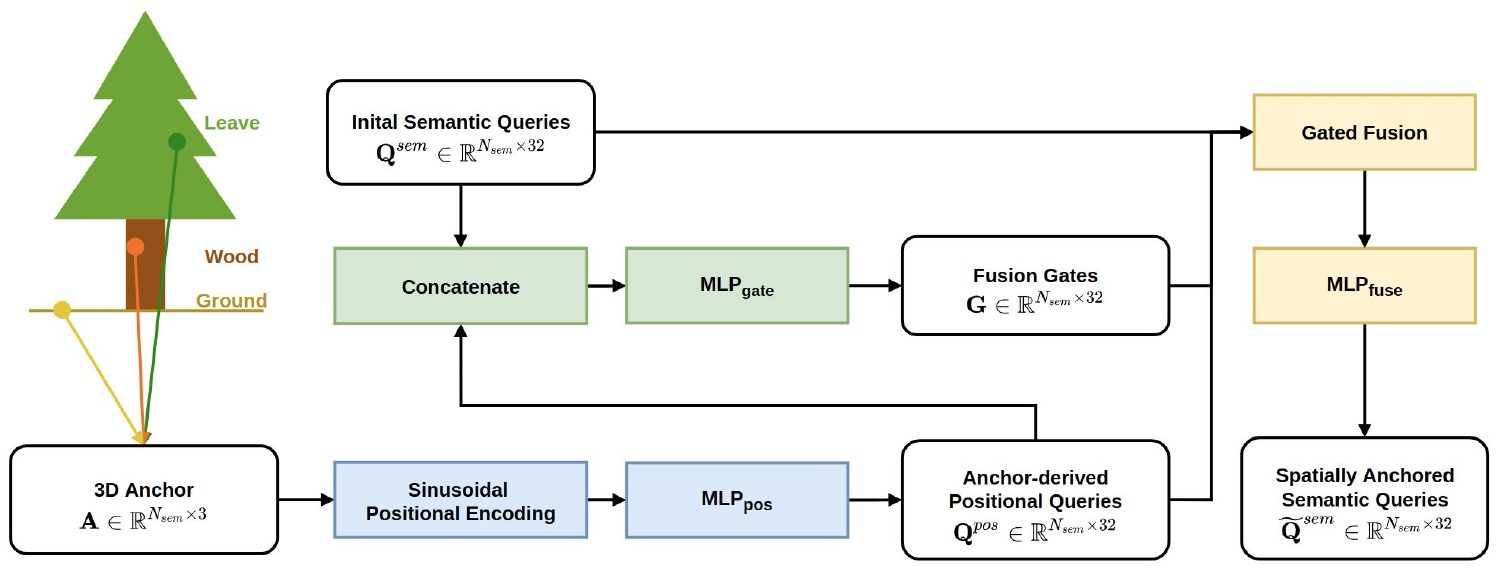}
    \caption{\textbf{Spatial anchoring and adaptive fusion for semantic query enhancement.}
    Learnable 3D anchors initialized with forest vertical stratification priors are encoded into positional queries and adaptively fused with learnable semantic queries through channel-wise gating, yielding spatially anchored semantic queries for subsequent decoding.}
    \label{fig:sa_sqe}
\end{figure*}

\subsubsection{ISA-Boundary-Guided Query Point Selection}

The proposed ISA-boundary-guided query point selection combines coarse foreground filtering with farthest point sampling (FPS)~\citep{qi2017pointnetplusplus} in a boundary-augmented discriminative feature space.
The coarse prediction $\mathbf{F}_2$ is first used to identify foreground tree voxels:
\begin{equation}
    \hat{y}_i^{\mathrm{bi}} =  \arg\max_{c} F_{2,i,c}.
\end{equation}
The corresponding foreground index set is
\begin{equation}
    \mathcal{I}_{\mathrm{fg}}  =
    \left\{
    i
    \;\middle|\;
    \hat{y}_i^{\mathrm{bi}}=1
    \right\}.
\end{equation}

For each candidate voxel $v_i$, $i\in\mathcal{I}_{\mathrm{fg}}$, let $\mathbf{e}_i\in\mathbb{R}^{5}$ denote its instance-discriminative feature from $\mathbf{F}_1$ and $b_i$ its predicted boundary probability.
The sampling representation is constructed as
\begin{equation}
    \mathbf{z}_i =
    \left[
    \mathbf{e}_i;
    \gamma_{\mathrm{bd}} b_i
    \right],
    \quad
    i\in\mathcal{I}_{\mathrm{fg}},
\end{equation}
where $\gamma_{\mathrm{bd}}$ controls the contribution of the boundary cue.
FPS is then performed in the augmented feature space:
\begin{equation}
    \mathcal{I}_{q}
    =
    \operatorname{FPS}
    \left(
    \left\{
    \mathbf{z}_i
    \mid
    i\in\mathcal{I}_{\mathrm{fg}}
    \right\},
    N_{\mathrm{ins}}
    \right),
\end{equation}
where $N_{\mathrm{ins}}$ denotes the number of instance query points and $\mathcal{I}_{q}$ contains the selected indices.
By augmenting the discriminative features with boundary probabilities, the sampling space jointly reflects instance separability and boundary variation.
The selected voxels are subsequently used to construct the instance queries $\mathbf{Q}^{\mathrm{ins}} \in\mathbb{R}^{N_{\mathrm{ins}}\times d}$.

\subsection{Spatially Anchored Semantic Query Enhancement}
\label{sec:method_sa_sqe}

As illustrated in Fig.~\ref{fig:sa_sqe}, we introduce SA-SQE to incorporate explicit spatial priors into semantic query representations.
SA-SQE associates each learnable semantic query with a learnable 3D anchor and adaptively integrates the resulting spatial representation with the semantic content.

Let $\mathbf{Q}^{\mathrm{sem}} = \left[\mathbf{q}^{\mathrm{sem}}_1, \ldots, \mathbf{q}^{\mathrm{sem}}_{N_{\mathrm{sem}}} \right]^{\top} \in \mathbb{R}^{N_{\mathrm{sem}}\times d}$ denote the learnable semantic queries, where $N_{\mathrm{sem}}$ is the number of semantic queries and $d=32$ matches the feature dimension of the shared feature representation $\mathbf{F}$.
Each semantic query is associated with a learnable 3D anchor, forming $\mathbf{A} = \left[\mathbf{a}_1, \ldots, \mathbf{a}_{N_{\mathrm{sem}}} \right]^{\top} \in \mathbb{R}^{N_{\mathrm{sem}}\times3}$.
The anchors are initialized according to the vertical stratification prior of forest components and are jointly optimized with the network during training.

The anchors are first encoded into positional queries through a 3D sinusoidal positional encoding followed by an MLP projection:
\begin{equation}
    \mathbf{Q}^{\mathrm{pos}}
    =
    \phi_{\mathrm{pos}}
    \left(
    \operatorname{PE}_{3\mathrm{D}}
    \left(
    \mathbf{A}
    \right)
    \right)
    \in
    \mathbb{R}^{N_{\mathrm{sem}}\times d},
\end{equation}
where $\phi_{\mathrm{pos}}(\cdot)$ denotes the positional projection MLP.

To adaptively balance semantic content and spatial information, a channel-wise fusion gate is predicted from the semantic and positional queries:
\begin{equation}
    \mathbf{G}
    =
    \sigma
    \left(
    \phi_{\mathrm{gate}}
    \left(
    \left[
    \mathbf{Q}^{\mathrm{sem}};
    \mathbf{Q}^{\mathrm{pos}}
    \right]
    \right)
    \right)
    \in
    \mathbb{R}^{N_{\mathrm{sem}}\times d},
\end{equation}
where $\phi_{\mathrm{gate}}(\cdot)$ denotes the learnable gating projection and $\sigma(\cdot)$ is the sigmoid function.
The two representations are then adaptively fused as
\begin{equation}
    \mathbf{R}
    =
    \mathbf{G}
    \odot
    \mathbf{Q}^{\mathrm{sem}}
    +
    \left(
    \mathbf{1}-\mathbf{G}
    \right)
    \odot
    \mathbf{Q}^{\mathrm{pos}},
\end{equation}
where $\odot$ denotes element-wise multiplication.
Finally, the spatially anchored semantic queries are obtained as
\begin{equation}
    \widetilde{\mathbf{Q}}^{\mathrm{sem}}
    =
    \phi_{\mathrm{fuse}}
    \left(
    \mathbf{R}
    \right)
    \in
    \mathbb{R}^{N_{\mathrm{sem}}\times d},
\end{equation}
where $\phi_{\mathrm{fuse}}(\cdot)$ denotes the final fusion network.
The resulting semantic queries are subsequently jointly decoded with the instance queries $\mathbf{Q}^{\mathrm{ins}} \in\mathbb{R}^{N_{\mathrm{ins}}\times d}$.

\subsection{Unified Query Decoding and Optimization}
\label{sec:method_decoder}

The constructed instance queries $\mathbf{Q}^{\mathrm{ins}}\in\mathbb{R}^{N_{\mathrm{ins}}\times d}$ and the spatially anchored semantic queries $\widetilde{\mathbf{Q}}^{\mathrm{sem}}\in\mathbb{R}^{N_{\mathrm{sem}}\times d}$ are jointly decoded with the shared feature representation $\mathbf{F}$.
Following the query decoder design of~\citep{kolodiazhnyi2024oneformer3d}, we employ an $L$-layer Transformer decoder to iteratively update the two query sets through interactions with the scene features:
\begin{equation}
    \left(
    \mathbf{Q}^{\mathrm{ins}}_{l},
    \mathbf{Q}^{\mathrm{sem}}_{l}
    \right)
    =
    \operatorname{Dec}^{(l)}
    \left(
    \mathbf{Q}^{\mathrm{ins}}_{l-1},
    \mathbf{Q}^{\mathrm{sem}}_{l-1},
    \mathbf{F}
    \right).
\end{equation}
Here, $l=1,\ldots,L$, with $\mathbf{Q}^{\mathrm{ins}}_{0}=\mathbf{Q}^{\mathrm{ins}}$ and $\mathbf{Q}^{\mathrm{sem}}_{0}=\widetilde{\mathbf{Q}}^{\mathrm{sem}}$.
The decoded instance and semantic queries are subsequently mapped to individual-tree instance masks and semantic masks, respectively.

During training, we follow the one-to-many query association strategy of ForestFormer3D~\citep{xiang2025forestformer3d}, where each ground-truth tree instance can be matched with multiple positive instance queries.
For the $g$-th ground-truth tree, let $\mathcal{I}_{g}^{+}$ denote the set of matched positive query indices and $N_g=|\mathcal{I}_{g}^{+}|$ the corresponding number of positive queries.
This association provides multiple supervised query hypotheses for each tree instance and enables query-specific reweighting according to boundary probability during subsequent optimization.

To account for the different reliability of these positive queries, we further use the predicted boundary probabilities to adaptively reweight their supervision.
For a positive query $i\in\mathcal{I}_{g}^{+}$, let $b_i^{q}$ denote the boundary probability at its sampled query location.
The corresponding supervision weight is defined as
\begin{equation}
    w_i
    =
    \frac{
        N_g
        \exp\left(-b_i^{q}/\tau\right)
    }{
        \displaystyle
        \sum_{j\in\mathcal{I}_{g}^{+}}
        \exp\left(-b_j^{q}/\tau\right)
    },
    \quad
    i\in\mathcal{I}_{g}^{+},
\end{equation}
where $\tau$ controls the sensitivity to boundary probability.
The factor $N_g$ normalizes the weights such that $\sum_{i\in\mathcal{I}_{g}^{+}} w_i=N_g$, while queries with higher boundary probabilities receive smaller relative weights.

The boundary-aware weights are subsequently applied to the supervision of the matched positive instance queries.
For decoder layer $l$, the loss associated with query $i\in\mathcal{I}_{g}^{+}$ is defined as
\begin{equation}
\begin{aligned}
    \ell_{i,g}^{(l)}
    =
    &\mathcal{L}_{\mathrm{bce}}
    \left(
        \hat{\mathbf{m}}_{i}^{(l)},
        \mathbf{m}_{g}^{\mathrm{gt}}
    \right)
    +
    \mathcal{L}_{\mathrm{dice}}
    \left(
        \hat{\mathbf{m}}_{i}^{(l)},
        \mathbf{m}_{g}^{\mathrm{gt}}
    \right)
    \\
    &+
    \lambda_{\mathrm{score}}
    \mathcal{L}_{\mathrm{score}}
    \left(
        \hat{s}_{i}^{(l)},
        \operatorname{IoU}
        \left(
            \hat{\mathbf{m}}_{i}^{(l)},
            \mathbf{m}_{g}^{\mathrm{gt}}
        \right)
    \right),
\end{aligned}
\end{equation}
where $\hat{\mathbf{m}}_{i}^{(l)}$ and $\hat{s}_{i}^{(l)}$ denote the predicted mask and confidence score of the $i$-th instance query, respectively, and $\mathbf{m}_{g}^{\mathrm{gt}}$ is the ground-truth mask of the $g$-th tree instance.
The boundary-aware instance loss at decoder layer $l$ is then
\begin{equation}
    \mathcal{L}_{\mathrm{ins}}^{(l)}
    =
    \sum_{g}
    \sum_{i\in\mathcal{I}_{g}^{+}}
    w_i
    \ell_{i,g}^{(l)}.
\end{equation}

The semantic loss $\mathcal{L}_{\mathrm{sem}}^{(l)}$ is computed using cross-entropy at decoder layer $l$, and supervision is applied to all decoder layers:
\begin{equation}
    \mathcal{L}_{\mathrm{dec}}
    =
    \sum_{l=1}^{L}
    \left(
        \mathcal{L}_{\mathrm{ins}}^{(l)}
        +
        \lambda_{\mathrm{sem}}
        \mathcal{L}_{\mathrm{sem}}^{(l)}
    \right),
\end{equation}
where $\lambda_{\mathrm{sem}}$ balances the instance and semantic objectives.

In addition to decoder supervision, the auxiliary predictions introduced in Section~\ref{sec:method_feature_query} are jointly optimized.
Following ForestFormer3D, we adopt $\mathcal{L}_{\mathrm{disc}}$ and $\mathcal{L}_{\mathrm{bi}}$ to supervise the instance-discriminative features $\mathbf{F}_{1}$ and the coarse tree/non-tree prediction $\mathbf{F}_{2}$, respectively.
The proposed boundary branch is additionally supervised by $\mathcal{L}_{\mathrm{bd}}$, as defined above.
The overall training objective is
\begin{equation}
    \mathcal{L}_{\mathrm{total}}
    =
    \mathcal{L}_{\mathrm{dec}}
    +
    \mathcal{L}_{\mathrm{disc}}
    +
    \mathcal{L}_{\mathrm{bi}}
    +
    \mathcal{L}_{\mathrm{bd}}.
\end{equation}

\section{Experiments}
\label{sec:exp}

\begin{table*}[t]
    \caption{\textbf{Summary of the datasets used in this study.} 
    FOR-instanceV2 is detailed by subset, with LAUTx and our self-collected dataset included for additional evaluation.}

    \label{tab:dataset_summary}
    \centering
    
    \small
    \setlength{\tabcolsep}{4pt}
    \renewcommand{\arraystretch}{1.15}

    \begin{tabularx}{\textwidth}{
        @{}
        >{\centering\arraybackslash}m{0.5cm}
        >{\raggedright\arraybackslash}m{1.9cm}
        >{\raggedright\arraybackslash}X
        >{\centering\arraybackslash}m{0.72cm}
        >{\centering\arraybackslash}m{0.72cm}
        >{\centering\arraybackslash}m{0.72cm}
        >{\centering\arraybackslash}m{1.2cm}
        >{\raggedright\arraybackslash}m{2.2cm}
        @{}
    }
        \toprule
        &
        \multirow{2}{*}{\textbf{Dataset}} &
        \multirow{2}{*}{\makecell[l]{\textbf{Scanning Mode}\\\textbf{(Sensor)}}} &
        \multicolumn{3}{c}{\textbf{Number of Plots}} &
        \multirow{2}{*}{\makecell[c]{\textbf{Number}\\\textbf{of Trees}}} &
        \multirow{2}{*}{\textbf{Country}} \\
        \cmidrule(lr){4-6}
        & & &
        \textbf{Train} &
        \textbf{Val} &
        \textbf{Test} &
        & \\
        \midrule

        \multirow{9}{*}{%
            \rotatebox[origin=c]{90}{%
                \textbf{FOR-instanceV2}%
            }%
        }
        & CULS
        & ULS (Riegl VUX-1)
        & 1 & 1 & 1
        & 37
        & Czech Republic \\

        & NIBIO
        & ULS (Riegl MiniVUX-1)
        & 8 & 6 & 6
        & 575
        & Norway \\

        & RMIT
        & ULS (Riegl MiniVUX-1)
        & 1 & 0 & 1
        & 223
        & Australia \\

        & SCION
        & ULS (Riegl MiniVUX-1)
        & 2 & 1 & 2
        & 135
        & New Zealand \\

        & TUWIEN
        & ULS (Riegl VUX-1)
        & 1 & 0 & 1
        & 150
        & Austria \\

        & NIBIO2
        & ULS (Riegl VUX-1)
        & 29 & 6 & 15
        & 3062
        & Norway \\

        & NIBIO\_MLS
        & MLS (GeoSLAM ZEB-HORIZON)
        & 4 & 1 & 1
        & 258
        & Norway \\

        & BlueCat
        & TLS (Leica P20)
        & 1 & 1 & 1
        & 6304
        & Czech Republic \\

        & Yuchen
        & ULS (Riegl MiniVUX-1)
        & 1 & 1 & 1
        & 281
        & French Guiana \\

        \midrule

        \multirow{2}{*}{%
            \rotatebox[origin=c]{90}{%
                \textbf{Extra}%
            }%
        }
        & LAUTx
        & MLS (GeoSLAM ZEB-HORIZON)
        & 0 & 0 & 6
        & 516
        & Austria \\

        & Self-collected
        & MLS (SHARE SLAM S20) \& ULS (RoboSense Airy) 
        & 0 & 0 & 12
        & 137
        & China \\

        \bottomrule
    \end{tabularx}
\end{table*}

\subsection{Experimental Setup}
\label{sec:exp_setup}

\subsubsection{Datasets}
We evaluate \ourmethod~on three forest point cloud datasets covering diverse sensing platforms, acquisition configurations, and scene conditions: FOR-instanceV2~\citep{xiang2025forestformer3d}, LAUTx~\citep{tockner2022lautx}, and our self-collected dataset.
Their main characteristics, including scanning modality, data split, number of annotated trees, and geographic region, are summarized in Table~\ref{tab:dataset_summary}.
FOR-instanceV2 is used for model training, validation, and benchmark evaluation, whereas LAUTx and the self-collected dataset are reserved for assessing generalization to unseen forest scenes and acquisition conditions.

FOR-instanceV2 comprises forest point clouds collected across multiple forest areas and sensing configurations, with point-level semantic annotations and individual-tree instance labels.
For semantic segmentation, each point is categorized as ground, wood, or leaf, while for instance segmentation, the wood and leaf points belonging to the same tree share a common instance identity and ground points are excluded from instance assignment.
Following the official split, the dataset contains 48 training scenes, 17 validation scenes, and 29 test scenes.

LAUTx contains six forest scenes acquired using mobile laser scanning (MLS) and serves as an external dataset for cross-dataset generalization evaluation.
We further construct a self-collected dataset comprising 12 test scenes acquired using UAV laser scanning (ULS) and MLS.
The self-collected scenes are acquired from geographic regions distinct from those represented in FOR-instanceV2 and LAUTx, introducing additional variations in tree species composition.
Together, these two held-out datasets differ from the training benchmark in observation viewpoint, spatial coverage, sampling characteristics, point density, occlusion patterns, and forest composition, providing complementary settings for evaluating the generalization of \ourmethod~to unseen forest scenes, sensing conditions, and tree species.

\subsubsection{Evaluation Metrics}
For individual-tree instance segmentation, predicted and ground-truth tree instances are matched according to their point-set intersection over union (IoU), with a threshold of 0.5.
Matched predictions are counted as true positives (TP), while unmatched predictions and ground-truth instances are counted as false positives (FP) and false negatives (FN), respectively.
We report precision (P), recall (R), and F1-score:
\begin{equation}
    \mathrm{P}=\frac{\mathrm{TP}}{\mathrm{TP}+\mathrm{FP}},\quad
    \mathrm{R}=\frac{\mathrm{TP}}{\mathrm{TP}+\mathrm{FN}},\quad
    \mathrm{F1}=\frac{2\mathrm{PR}}{\mathrm{P}+\mathrm{R}}.
\end{equation}

Following the evaluation protocol of ForestFormer3D~\citep{xiang2025forestformer3d}, we also report coverage (Cov) to quantify instance delineation quality:
\begin{equation}
    \mathrm{Cov}
    =
    \frac{1}{G}
    \sum_{g=1}^{G}
    \max_{k}
    \operatorname{IoU}
    \left(
    \hat{\mathbf{m}}_{k},
    \mathbf{m}^{\mathrm{gt}}_{g}
    \right),
\end{equation}
where $G$ denotes the number of ground-truth tree instances, and $\hat{\mathbf{m}}_{k}$ and $\mathbf{m}^{\mathrm{gt}}_{g}$ denote the $k$-th predicted instance mask and the $g$-th ground-truth instance mask, respectively.

For semantic segmentation, we report the class-wise IoU and mean IoU (mIoU):
\begin{equation}
    \mathrm{IoU}_{c}
    =
    \frac{\mathrm{TP}_{c}}
    {\mathrm{TP}_{c}+\mathrm{FP}_{c}+\mathrm{FN}_{c}},
    \quad
    \mathrm{mIoU}
    =
    \frac{1}{C}
    \sum_{c=1}^{C}\mathrm{IoU}_{c},
\end{equation}
where $C$ is the number of semantic categories.

\subsubsection{Implementation Details}
The network takes only 3D point coordinates as input, without using additional color or intensity attributes.
During training, cylindrical regions with a radius of 16~m are randomly cropped from individual forest scenes and voxelized at a resolution of 0.2~m.
Data augmentation includes random horizontal flipping, rotation around the vertical axis within $[-\pi,\pi]$, scaling within $[0.8,1.2]$, and Gaussian translation with a standard deviation of 0.1~m.

For boundary target construction, the upper-canopy threshold ratio and relative boundary width are set to $\eta=0.25$ and $\alpha=0.2$, respectively.
The BEV grid resolution is set to $s_v=0.2$~m, consistent with the voxel resolution.
The 3D contact boundary is constructed within a local neighborhood of radius $r_{\mathrm{c}}=0.5$~m, and the positive-class weight in the boundary loss is set to $w_{\mathrm{pos}}=5.0$.

For ISA-boundary-guided query point selection, the boundary scaling factor is set to $\gamma_{\mathrm{bd}}=2$, and the number of instance queries is $N_{\mathrm{ins}}=300$.
For SA-SQE, we use $N_{\mathrm{sem}}=3$ semantic queries.
The corresponding learnable anchors are initialized with zero horizontal coordinates and vertical positions of $0$, $3$, and $8$~m for ground, wood, and leaf, respectively, and are jointly optimized during training.

The Transformer decoder contains $L=6$ layers.
The BCE and Dice losses for instance segmentation are assigned unit weights, while $\lambda_{\mathrm{score}}=0.5$ and $\lambda_{\mathrm{sem}}=0.2$.
For adaptive boundary-aware reweighting, the temperature parameter is set to $\tau=1.0$.
The network is optimized using AdamW with an initial learning rate of $2\times10^{-4}$, a weight decay of 0.05, and gradient clipping with a maximum norm of 10.
A PolyLR schedule with a power of 0.9 is adopted for learning-rate decay.
Training is performed for 450,000 iterations on a single NVIDIA RTX PRO 6000 GPU with a batch size of 2.

At inference, each scene is processed using overlapping cylindrical windows with a radius of 16~m and a stride of 4~m.
Predictions whose mask centroids lie within 0.5~m of the cylindrical boundary are discarded, and the remaining predictions are transformed to the original point-cloud coordinate system.
Following ForestFormer3D, instance masks are ranked by decoder confidence scores and merged using score-based non-maximum suppression (NMS), while semantic predictions are aggregated by majority voting.
Finally, instance assignments on ground points are removed, and instances containing fewer than 10 points are filtered out.

\subsection{Comparison with State-of-the-Art Methods}
\label{sec:exp_sota}

\begin{table*}[t]
    \centering
    \caption{\textbf{Quantitative comparison with state-of-the-art methods on the FOR-instanceV2 test set.}
    \textbf{Bold} and \underline{underlined} values denote the best and second-best results, respectively.
    $^{*}$TreeLearn results exclude the Yuchen region.}
    \label{tab:comparison_forinstance}

    \small
    \setlength{\tabcolsep}{8pt}
    \renewcommand{\arraystretch}{1.15}

    \begin{tabular}{lcccccccc}
        \toprule
        \multirow{2}{*}{\textbf{Method}} &
        \multicolumn{4}{c}{\textbf{Individual Tree Seg. (\%)}} &
        \multicolumn{4}{c}{\textbf{Semantic Seg. (\%)}} \\
        \cmidrule(lr){2-5}
        \cmidrule(lr){6-9}
        &
        \textbf{P} &
        \textbf{R} &
        \textbf{F1} &
        \textbf{Cov} &
        \textbf{Ground} &
        \textbf{Wood} &
        \textbf{Leaf} &
        \textbf{mIoU} \\
        \midrule
        
        ForAINetV2\_R16
        & 88.1 & 59.2 & 70.8 & 72.7
        & \underline{98.8} & \underline{69.2} & \textbf{94.5} & \underline{87.5} \\

        TreeLearn$^{*}$
        & 82.0 & 36.6 & 50.6 & 52.2
        & -- & -- & -- & -- \\

        OneFormer3D
        & 72.0 & 74.3 & 73.1 & 80.0
        & 97.8 & 63.1 & 93.4 & 84.8 \\

        ForestFormer3D
        & \textbf{92.4} & \underline{75.0} & \underline{82.8} & \underline{81.2}
        & 96.9 & 67.7 & 94.1 & 86.2 \\

        \ourmethod~(ours)
        & \underline{91.8} & \textbf{77.3} & \textbf{84.0} & \textbf{90.0}
        & \textbf{99.0} & \textbf{69.6} & \underline{94.3} & \textbf{87.6} \\
        
        \bottomrule
    \end{tabular}
\end{table*}

We compare \ourmethod~with representative state-of-the-art (SOTA) methods selected according to relevance, recency, and reproducibility, prioritizing approaches with publicly available implementations or sufficiently complete experimental configurations.
The baselines include forest-specific methods ForAINetV2~\citep{xiang2024automated} and TreeLearn~\citep{henrich2024treelearn}, the general unified segmentation framework OneFormer3D~\citep{kolodiazhnyi2024oneformer3d}, and the recent query-based forest segmentation method ForestFormer3D~\citep{xiang2025forestformer3d}.

\subsubsection{Quantitative Results}

Table~\ref{tab:comparison_forinstance} reports the quantitative results on the FOR-instanceV2 test set.
TreeLearn fails to predict individual trees in the Yuchen region due to the extreme sparsity of stem points; its reported results therefore exclude this region.

For individual-tree segmentation, \ourmethod~achieves the best F1-score of 84.0\% and coverage of 90.0\%.
Compared with ForestFormer3D, the F1-score and coverage are improved by 1.2 and 8.8 percentage points, respectively.
Although ForestFormer3D attains a slightly higher precision (92.4\% vs.~91.8\%), \ourmethod~increases recall from 75.0\% to 77.3\%, yielding a more favorable precision--recall balance.
The advantage is more pronounced in coverage, where \ourmethod~substantially outperforms all competing methods.
This indicates that its predicted instances not only recover a larger fraction of ground-truth trees, but also achieve stronger spatial overlap with the corresponding instances.
In contrast, ForAINetV2\_R16 and TreeLearn exhibit considerably lower recall and coverage, while OneFormer3D sacrifices precision for higher recall, highlighting the difficulty of simultaneously maintaining detection accuracy and instance completeness.

For semantic segmentation, \ourmethod~achieves the best ground and wood IoUs of 99.0\% and 69.6\%, respectively, together with the highest mIoU of 87.6\%.
Its leaf IoU reaches 94.3\%, only 0.2 percentage points below the best result of ForAINetV2\_R16.
Compared with ForestFormer3D, \ourmethod~improves ground, wood, and mIoU by 2.1, 1.9, and 1.4 percentage points, respectively, while maintaining competitive performance on leaf.
The class-wise results also show that wood remains substantially more challenging than ground and leaf across all methods, suggesting that semantic ambiguity is concentrated mainly in woody structures rather than being evenly distributed across classes.
Overall, \ourmethod~achieves a more balanced improvement across both individual-tree and semantic segmentation, without trading a substantial loss in either task.

\subsubsection{Qualitative Results}

\begin{figure*}[htbp]
    \centering
    \includegraphics[width=0.95\linewidth]{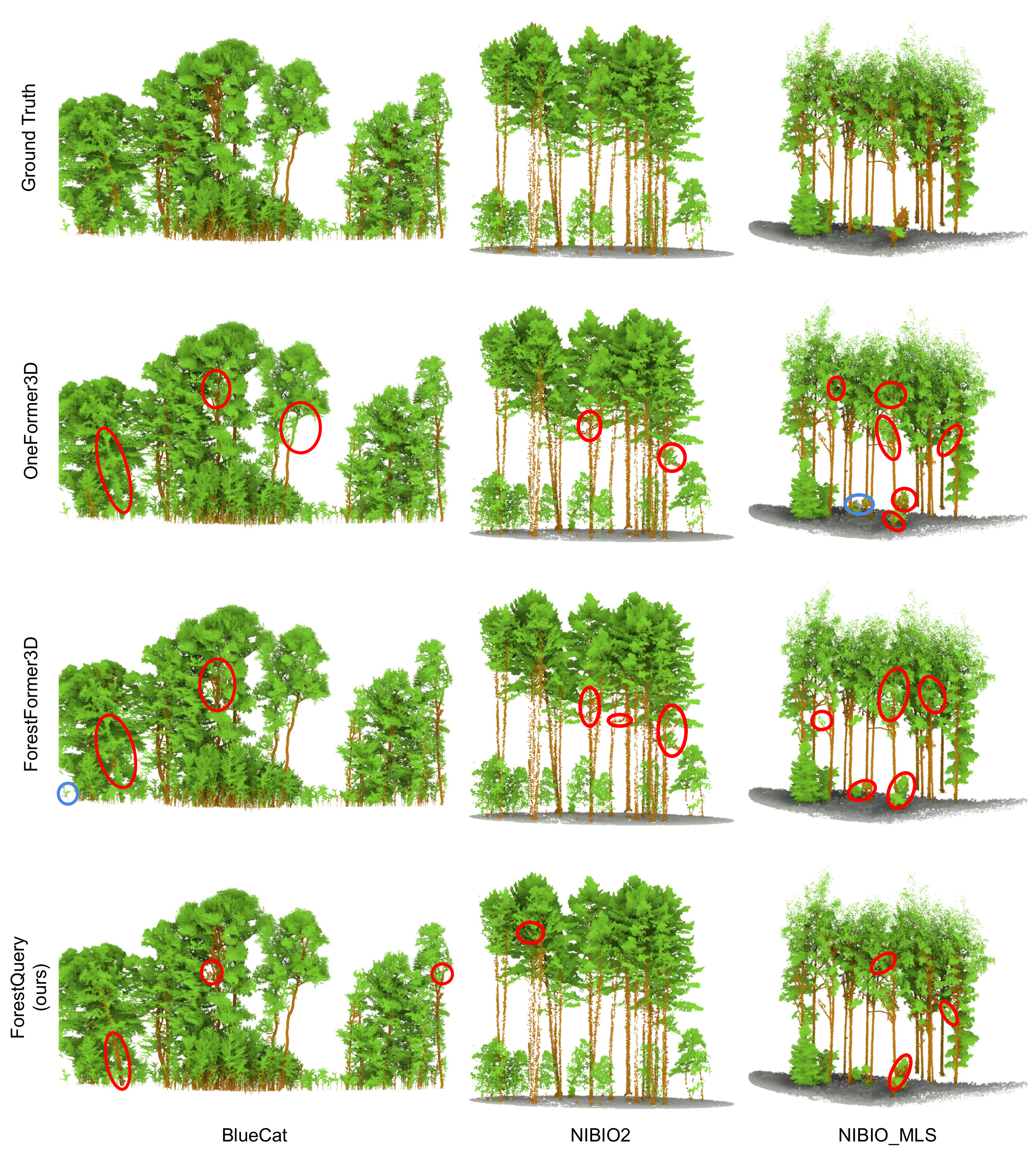}
    \caption{\textbf{Qualitative comparison of semantic segmentation on representative scenes from the BlueCat, NIBIO2, and NIBIO\_MLS subsets of the FOR-instanceV2 test set.} 
    Leaf, wood, and ground points are shown in green, brown, and gray, respectively. 
    Red circles highlight confusion between wood and leaf points, while blue circles indicate tree points incorrectly classified as ground.}
    
    \label{fig:semantic_forinstancev2}
\end{figure*}

\begin{figure*}[htbp]
    \centering
    \includegraphics[width=0.95\linewidth]{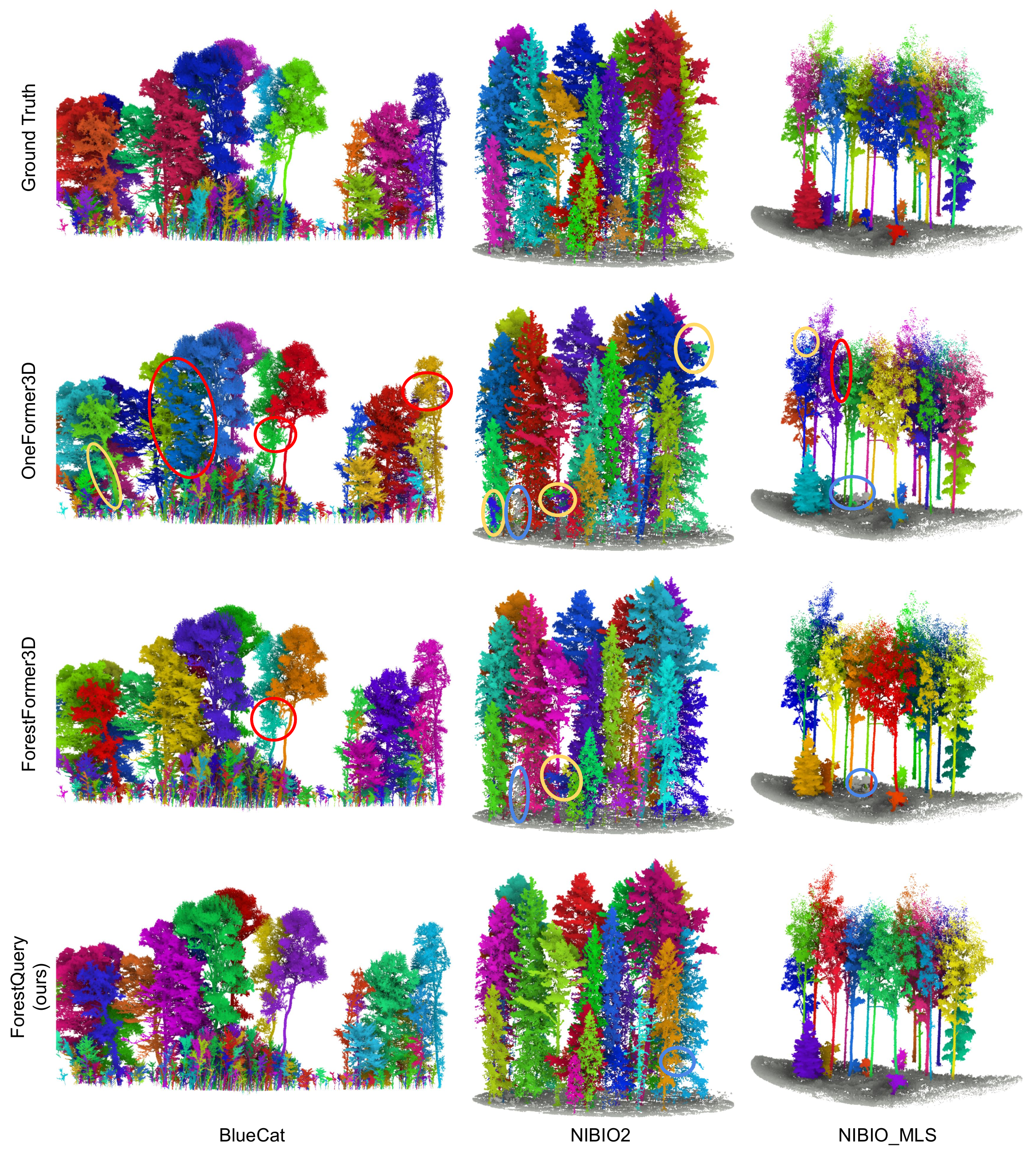}
    \caption{\textbf{Qualitative comparison of individual-tree segmentation on representative scenes from the BlueCat, NIBIO2, and NIBIO\_MLS subsets of the FOR-instanceV2 test set.} 
    Ground points are shown in gray, while different colors denote individual-tree instances. 
    Yellow circles highlight over-segmentation, blue circles indicate undetected trees, and red circles mark points incorrectly assigned from one tree to a neighboring tree.}

    \label{fig:instance_forinstancev2}
\end{figure*}

To complement the quantitative evaluation, qualitative comparisons focus on OneFormer3D and ForestFormer3D, the most directly comparable query-based unified segmentation baselines.
TreeLearn does not provide multi-class semantic predictions, while ForAINetV2 adopts a non-query-based formulation; both are retained in the quantitative comparison.
Representative scenes are selected from the BlueCat, NIBIO2, and NIBIO\_MLS subsets of the FOR-instanceV2 test set to cover diverse sensing characteristics and forest structures.

Fig.~\ref{fig:semantic_forinstancev2} compares the semantic segmentation results.
OneFormer3D and ForestFormer3D exhibit noticeable confusion between wood and leaf components, particularly in structurally complex canopy regions where branches and foliage are spatially intertwined.
Tree points are also occasionally misclassified as ground near the lower parts of the scenes, indicating difficulty in maintaining semantic consistency across tree--ground transitions.
In comparison, \ourmethod~produces cleaner and more spatially coherent semantic predictions across all three subsets, with fewer wood--leaf confusions and tree-to-ground errors.
The improvement is particularly visible in regions with dense canopy structure and around tree bases, where class boundaries are less geometrically distinct.
These observations are consistent with the higher wood IoU and overall mIoU reported in Table~\ref{tab:comparison_forinstance}, while the already high leaf IoU across methods suggests that the remaining semantic difficulty is concentrated mainly in woody and transition regions.

Fig.~\ref{fig:instance_forinstancev2} presents the corresponding individual-tree segmentation results.
Yellow, blue, and red circles denote over-segmentation, undetected trees, and incorrect point assignment across neighboring instances, respectively.
OneFormer3D exhibits all three types of errors, reflecting difficulties in simultaneously preserving complete tree instances and separating neighboring trees.
ForestFormer3D reduces several of these errors, but inaccurate separation and cross-instance assignment remain visible in regions with dense tree distributions and overlapping crowns.
By comparison, \ourmethod~produces more coherent tree instances, with fewer missed trees, fragmented instances, and erroneous assignments between adjacent trees.
The improvement is consistently observed across the three subsets rather than being confined to a particular scene configuration.
This behavior agrees with the quantitative results, where \ourmethod~achieves both the highest F1-score and a substantially higher coverage, indicating improved instance recognition together with more complete tree delineation.

\subsection{Ablation Studies}
\label{sec:exp_abl}

\begin{table*}[t]
    \centering
    \caption{\textbf{Ablation study on the FOR-instanceV2 test set.}
    \textbf{Bold} and \underline{underlined} values denote the best and second-best results, respectively.}
    \label{tab:ablation}

    \small
    \setlength{\tabcolsep}{8pt}
    \renewcommand{\arraystretch}{1.2}

    \begin{tabular}{ccccccccccc}
        \toprule
        \multirow{2}{*}{\textbf{Baseline}} &
        \multirow{2}{*}{\textbf{BU}} &
        \multirow{2}{*}{\textbf{SA-SQE}} &
        \multicolumn{4}{c}{\textbf{Individual Tree Seg. (\%)}} &
        \multicolumn{4}{c}{\textbf{Semantic Seg. (\%)}} \\
        \cmidrule(lr){4-7}
        \cmidrule(lr){8-11}
        & & &
        \textbf{P} &
        \textbf{R} &
        \textbf{F1} &
        \textbf{Cov} &
        \textbf{Ground} &
        \textbf{Wood} &
        \textbf{Leaf} &
        \textbf{mIoU} \\
        \midrule

        \checkmark &  & 
        & \textbf{92.4} & 75.0 & 82.8 & 81.2
        & 96.9 & 67.7 & \underline{94.1} & 86.2 \\

        \checkmark & \checkmark & 
        & 91.7 & \textbf{77.9} & \textbf{84.2} & \textbf{90.1}
        & \underline{98.5} & 66.9 & 94.0 & 86.5 \\

        \checkmark &  & \checkmark
        & 90.9 & 76.7 & 83.2 & \textbf{90.1}
        & \textbf{99.0} & \underline{68.0} & \underline{94.1} & \underline{87.0} \\

        \checkmark & \checkmark & \checkmark
        & \underline{91.8} & \underline{77.3} & \underline{84.0} & \underline{90.0}
        & \textbf{99.0} & \textbf{69.6} & \textbf{94.3} & \textbf{87.6} \\

        \bottomrule
    \end{tabular}
\end{table*}

We conduct component-wise ablations on the FOR-instanceV2 test set to examine the contributions of boundary uncertainty (BU) and SA-SQE.
ForestFormer3D serves as the common baseline, providing a query-based unified segmentation framework for controlled evaluation of the proposed designs.
BU and SA-SQE are introduced individually and jointly, allowing their task-specific contributions and joint effect to be assessed under the same experimental setting.
Table~\ref{tab:ablation} reports both individual-tree and semantic segmentation metrics.

\textbf{Effect of BU.}
Introducing BU leads to clear improvements in individual-tree segmentation.
Compared with the baseline, recall increases from 75.0\% to 77.9\%, F1-score from 82.8\% to 84.2\%, and coverage from 81.2\% to 90.1\%, while precision decreases only slightly from 92.4\% to 91.7\%.
The substantially larger gain in coverage indicates that BU is particularly effective in improving the spatial completeness of predicted tree instances, in addition to increasing instance-level recognition performance.
Its effect on semantic segmentation is comparatively limited: mIoU increases from 86.2\% to 86.5\%, with an improvement in ground IoU but minor decreases in wood and leaf IoUs.
This performance pattern is consistent with the design objective of BU, which focuses on ambiguous instance boundaries and individual-tree delineation.

\textbf{Effect of SA-SQE.}
SA-SQE exhibits a different performance profile, with its primary improvement appearing in semantic segmentation.
It improves ground IoU from 96.9\% to 99.0\% and wood IoU from 67.7\% to 68.0\%, while preserving the leaf IoU at 94.1\%, resulting in an mIoU increase from 86.2\% to 87.0\%.
The semantic gain is driven mainly by the improved ground prediction, together with a moderate improvement in the more challenging wood class, without degrading leaf segmentation.
SA-SQE also benefits individual-tree segmentation: recall and F1-score increase to 76.7\% and 83.2\%, respectively, while coverage rises markedly to 90.1\%.
This cross-task improvement suggests that spatially structured semantic queries can provide useful information to the shared unified segmentation process beyond semantic prediction alone.

\textbf{Joint Effect of BU and SA-SQE.}
Combining BU and SA-SQE provides the strongest overall balance between individual-tree and semantic segmentation.
Relative to the baseline, the full model improves recall, F1-score, and coverage by 2.3, 1.2, and 8.8 percentage points, respectively, while improving ground, wood, and leaf IoUs by 2.1, 1.9, and 0.2 percentage points and mIoU by 1.4 percentage points.
Although BU alone achieves slightly higher F1-score and coverage (84.2\% and 90.1\% versus 84.0\% and 90.0\%), incorporating SA-SQE largely preserves these instance-level gains while increasing wood IoU from 66.9\% to 69.6\% and mIoU from 86.5\% to 87.6\%.
Conversely, adding BU to SA-SQE further improves precision, recall, and F1-score by 0.9, 0.6, and 0.8 percentage points, respectively, while maintaining nearly identical coverage and further increasing wood IoU and mIoU by 1.6 and 0.6 percentage points.
These two directions of improvement show that BU and SA-SQE address complementary aspects of the unified task: the former primarily strengthens individual-tree delineation, whereas the latter enhances semantic discrimination.
Their combination therefore yields a more balanced improvement across both instance and semantic segmentation.

\subsection{Generalization Evaluation}
\label{sec:exp_general}

\begin{table*}[t]
    \centering
    \caption{\textbf{Quantitative comparison of generalization performance on the LAUTx and self-collected datasets.}
    \textbf{Bold} and \underline{underlined} values denote the best and second-best results, respectively.}
    \label{tab:comparison_generalization}

    \small
    \setlength{\tabcolsep}{8pt}
    \renewcommand{\arraystretch}{1.15}

    \begin{tabular}{llccccccc}
        \toprule
        \multirow{2}{*}{\textbf{Dataset}} &
        \multirow{2}{*}{\textbf{Method}} &
        \multicolumn{4}{c}{\textbf{Individual Tree Seg. (\%)}} &
        \multicolumn{3}{c}{\textbf{Semantic Seg. (\%)}} \\
        \cmidrule(lr){3-6}
        \cmidrule(lr){7-9}
        & &
        \textbf{P} &
        \textbf{R} &
        \textbf{F1} &
        \textbf{Cov} &
        \textbf{Ground} &
        \textbf{Tree} &
        \textbf{mIoU} \\
        \midrule

        \multirow{5}{*}{\textbf{LAUTx}}
        & ForAINetV2\_R16
        & 93.7 & 77.7 & 85.0 & 74.7
        & \textbf{95.9} & \underline{96.7} & \textbf{96.3} \\

        & TreeLearn
        & \textbf{96.8} & \underline{87.9} & \textbf{92.1} & \underline{85.2}
        & 92.6 & 95.9 & 94.3 \\

        & OneFormer3D
        & 58.1 & 80.2 & 67.4 & 74.6
        & 94.6 & \textbf{96.9} & 95.8 \\

        & ForestFormer3D
        & \underline{95.7} & 85.9 & \underline{90.5} & 79.1
        & 94.1 & \underline{96.7} & 95.4 \\

        & \ourmethod~(ours)
        & 92.5 & \textbf{88.6} & \underline{90.5} & \textbf{86.8}
        & \underline{95.1} & \underline{96.7} & \underline{95.9} \\

        \midrule

        \multirow{5}{*}{\textbf{Self-collected}}
        & ForAINetV2\_R16
        & 78.3 & 86.9 & 82.4 & 84.3
        & 93.5 & 99.0 & 96.2 \\

        & TreeLearn
        & 38.4 & 33.3 & 35.7 & 47.2
        & 70.3 & 92.8 & 81.6 \\

        & OneFormer3D
        & 69.3 & 77.4 & 73.1 & 74.2
        & 57.5 & 93.8 & 75.6 \\

        & ForestFormer3D
        & \underline{90.2}
        & \underline{87.6}
        & \underline{88.9}
        & \underline{86.8}
        & \underline{97.3}
        & \underline{99.5}
        & \underline{98.4} \\

        & \ourmethod~(ours)
        & \textbf{97.7}
        & \textbf{93.4}
        & \textbf{95.5}
        & \textbf{90.7}
        & \textbf{98.2}
        & \textbf{99.7}
        & \textbf{99.0} \\

        \bottomrule
    \end{tabular}
\end{table*}

We further evaluate \ourmethod~on LAUTx and the self-collected dataset, neither of which is used during training.
Compared with FOR-instanceV2, both datasets introduce substantial domain shifts in sensing conditions and forest structure, while the self-collected scenes further differ in geographic region and tree species composition.
Both datasets provide individual-tree annotations and binary semantic labels for ground and tree, enabling evaluation of individual-tree segmentation and binary semantic generalization under unseen forest environments.
For semantic evaluation, the predicted wood and leaf classes are merged into the tree class, yielding two semantic categories ($C=2$).

\subsubsection{Quantitative Results}

Table~\ref{tab:comparison_generalization} reports the quantitative results on the two held-out generalization datasets.
On LAUTx, \ourmethod~achieves the highest recall of 88.6\% and coverage of 86.8\%, outperforming the second-best results by 0.7 and 1.6 percentage points, respectively.
Its F1-score reaches 90.5\%, tying with ForestFormer3D for the second-best result, although TreeLearn attains higher precision and the highest F1-score.
This performance profile indicates that \ourmethod~favors more complete tree recovery and delineation under the domain shift, as reflected by its leading recall and coverage, while maintaining competitive overall instance accuracy.
For binary semantic segmentation, \ourmethod~achieves ground and tree IoUs of 95.1\% and 96.7\%, respectively, yielding an mIoU of 95.9\%.
This is the second-best semantic result on LAUTx and is only 0.4 percentage points below the highest mIoU of 96.3\%, showing that the instance-level gains are obtained without a substantial degradation in tree--ground discrimination.

More pronounced gains are observed on the self-collected dataset, which introduces additional variations in acquisition conditions, geographic regions, and tree species composition.
\ourmethod~achieves the best results across all reported metrics, reaching 97.7\% precision, 93.4\% recall, 95.5\% F1-score, and 90.7\% coverage for individual-tree segmentation.
Compared with ForestFormer3D, the strongest competing query-based baseline, these results correspond to improvements of 7.5, 5.8, 6.6, and 3.9 percentage points, respectively.
The simultaneous gains in precision and recall are particularly notable, suggesting that the improvement is not obtained by simply favoring either conservative or aggressive instance prediction.
Instead, \ourmethod~maintains both accurate tree recognition and more complete instance delineation under the unseen acquisition and forest conditions.
For binary semantic segmentation, it further achieves 98.2\% ground IoU, 99.7\% tree IoU, and 99.0\% mIoU, improving over ForestFormer3D by 0.9, 0.2, and 0.6 percentage points, respectively.
Taken together, the results across LAUTx and the self-collected dataset show that \ourmethod~preserves strong semantic discrimination while exhibiting particularly consistent cross-dataset generalization for individual-tree segmentation.

\subsubsection{Qualitative Results}

\begin{figure*}[htbp]
    \centering
    \includegraphics[width=0.95\linewidth]{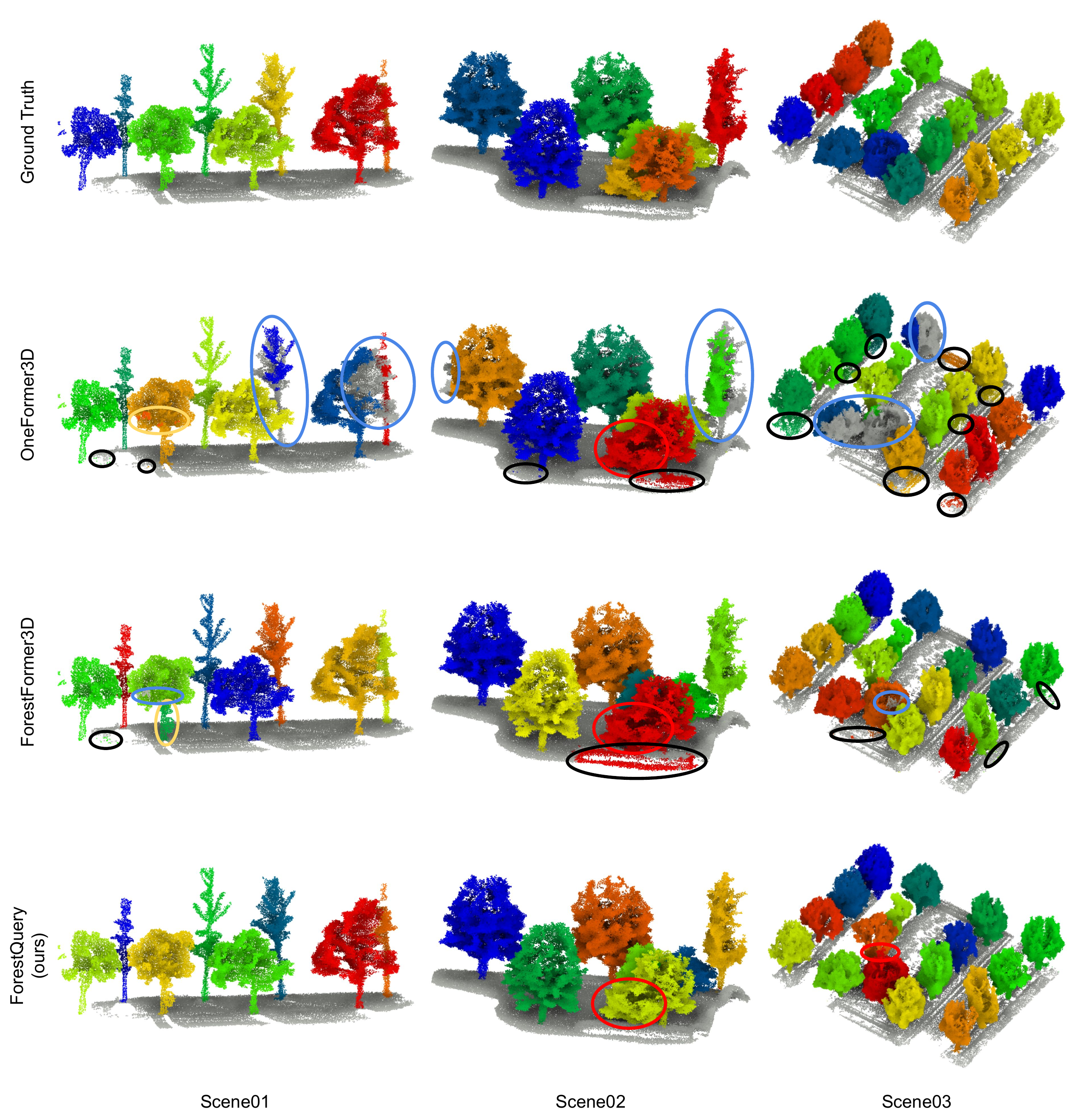}
    \caption{\textbf{Qualitative comparison of individual-tree segmentation on the generalization datasets.} 
    Ground points are shown in gray, while different colors denote individual-tree instances. 
    Yellow circles highlight over-segmentation, blue circles indicate undetected trees, red circles mark points incorrectly assigned from one tree to a neighboring tree, and black circles indicate ground points incorrectly assigned to tree instances.}

    \label{fig:instance_test}
\end{figure*}

Qualitative evaluation focuses on individual-tree segmentation to maintain a consistent visualization protocol across the two held-out datasets.
Since both LAUTx and the self-collected dataset provide only binary ground/tree semantic annotations, we omit separate semantic visualizations and present instance segmentation results instead.
Because tree points are assigned to individual instances while ground remains unassigned, these results jointly reflect individual-tree delineation and tree--ground separation.

Fig.~\ref{fig:instance_test} compares the predictions across three representative held-out scenes with different tree distributions and structural configurations.
OneFormer3D exhibits several recurring failure modes, including missed trees, over-segmentation, cross-instance assignment, and ground leakage into predicted tree instances.
These errors are particularly evident around slender trees, closely spaced individuals, and ambiguous tree--ground transition regions.
ForestFormer3D produces more coherent instances and reduces many of these errors, but residual failures remain near tree bases and between adjacent trees, where local geometric ambiguity makes instance separation difficult.

In comparison, \ourmethod~produces more complete and spatially coherent tree instances across the three scenes, with fewer missed trees, fragmented predictions, cross-instance assignments, and ground leakage.
The remaining errors are mainly confined to locally complex regions involving closely interacting neighboring trees.
Overall, the qualitative results indicate improved robustness to unseen scene structures while preserving both instance completeness and tree--ground consistency.
This observation is consistent with the quantitative results in Table~\ref{tab:comparison_generalization}, where \ourmethod~shows particularly strong recall and coverage across the held-out evaluation settings.

\section{Conclusion and Future Work}
\label{sec:conclusion}

In this work, we presented ForestQuery, a unified query-based framework for semantic and individual-tree segmentation of forest point clouds.
ForestQuery introduces a boundary-aware mechanism for instance query learning and SA-SQE for incorporating forest-specific spatial priors into semantic queries.
Experiments on FOR-instanceV2 demonstrate improved individual-tree delineation and semantic segmentation over representative state-of-the-art methods, while the ablation studies verify the complementary effects of BU and SA-SQE.
Evaluations on LAUTx and the self-collected dataset further show promising generalization to unseen sensing conditions and forest environments.

Future work will explore both methodological and data-centric extensions.
More adaptive spatial priors and stronger cross-domain modeling may further improve robustness to diverse forest structures and sensing conditions.
Meanwhile, larger and more diverse forest point cloud datasets could support large-scale pretraining toward generalizable forest point cloud foundation models, enabling transferable representations across species, regions, sensing platforms, and downstream forest understanding tasks.



\printcredits

\bibliographystyle{cas-model2-names}

\bibliography{refs}





\end{document}